\documentclass[letterpaper, 10 pt, journal, twoside]{IEEEtran}

\IEEEoverridecommandlockouts % This command is only needed if you want to use the \thanks command

\usepackage[utf8]{inputenc}
\usepackage{array, makecell} %
\usepackage{algorithm}
\usepackage[noend]{algpseudocode}
\usepackage{algpseudocode}
\usepackage{amsmath}
\usepackage{amssymb}
\usepackage{booktabs}
\usepackage[font=footnotesize]{caption}
\usepackage{cite}
\usepackage{color}
\usepackage{comment}
\usepackage{dblfloatfix}
\usepackage{graphicx}
\usepackage{epsfig}
\usepackage{epstopdf}
\usepackage[hidelinks]{hyperref}
\usepackage{cleveref}
\usepackage{inputenc}
\usepackage{lipsum}
\usepackage{multicol}
\usepackage{multirow}
\usepackage{setspace}
\usepackage{soul}
\usepackage{tabularx}
\usepackage{todonotes}
\usepackage{url}
\usepackage{varwidth}
\DeclareMathOperator*{\argmax}{arg\,max}
\makeatletter
\def\BState{\State\hskip-\ALG@thistlm}
\makeatother

\makeatletter
\newcommand{\algrule}[1][.2pt]{\par\vskip.5\baselineskip\hrule height #1\par\vskip.5\baselineskip}
\makeatother

\providecommand{\keywords}[1]{\textbf{\textit{Index Terms---}} #1}
\title{Learning to Communicate Functional States with Nonverbal Expressions for Improved Human-Robot Collaboration}

\author{Liam Roy$^{1}$, Elizabeth A. Croft$^{2}$, and Dana Kuli{\'c}$^{3}$ % <-this % stops a space
\thanks{$^{1}$Liam Roy is with Monash University
{\tt\footnotesize \href{mailto:liam.roy@monash.edu}{liam.roy@monash.edu}}}%
\thanks{$^{2}$Elizabeth A. Croft is with University of Victoria
{\tt\footnotesize \href{mailto:ecroft@uvic.ca}{ecroft@uvic.ca} }}%
\thanks{$^{3}$Dana Kuli{\'c} is with Monash University
{\tt\footnotesize \href{mailto:dana.kulic@monash.edu}{dana.kulic@monash.edu}}}
\thanks{D. Kuli{\'c} is supported by the ARC Future Fellowship (FT200100761).}
\thanks{Received: Nov 06 2023; Revised Feb 05 2024; Accepted Mar 03 2024}
\thanks{This paper was recommended for publication by Editor Gentiane Venture upon evaluation of the Associate Editor and Reviewers' comments.}
\thanks{Digital Object Identifier (DOI): see top of this page.}
}

\begin{document}

\maketitle

% \thispagestyle{plain}   % Put this right after make-title to force page numbers on IEEE
% \pagestyle{plain}       % Put this right after make-title to force page numbers on IEEE

% to highlight text: \hl{smooth interaction between}

\begin{abstract}
    Collaborative robots must effectively communicate their internal state to humans to enable a smooth interaction. Nonverbal communication is widely used to communicate information during human-robot interaction, however, such methods may also be misunderstood, leading to communication errors. In this work, we explore modulating the acoustic parameter values (pitch bend, beats per minute, beats per loop) of nonverbal auditory expressions to convey functional robot states (accomplished, progressing, stuck). We propose a reinforcement learning (RL) algorithm based on noisy human feedback to produce accurately interpreted nonverbal auditory expressions. The proposed approach was evaluated through a user study with 24 participants. The results demonstrate that: (i) Our proposed RL-based approach is able to learn suitable acoustic parameter values which improve the users’ ability to correctly identify the state of the robot. (ii) Algorithm initialization informed by previous user data can be used to significantly speed up the learning process. (iii) The method used for algorithm initialization strongly influences whether participants converge to similar sounds for each robot state. (iv) Modulation of pitch bend has the largest influence on user association between sounds and robotic states.
\end{abstract}


\section{Introduction}\label{sec:introduction}
% 3/4 of a page. Include 1 figure which summarizes work.
\IEEEPARstart{T}o facilitate seamless human-robot interaction (HRI), collaborative robots require the ability to convey their internal state to human teammates. Nonverbal communication forms an essential component of human interactions and has been an important focus in the development of HRIs \cite{Breazeal2005EffectsTeamwork, Knight2016LabanLanguage, Venture2019RobotMotions}. Humans frequently communicate with nonverbal gestures, and in order to achieve a comparable degree of fluidity during collaboration, robots may also benefit from utilizing nonverbal communication strategies. Nonverbal expressions including body language \cite{Knight2016LabanLanguage}, gestures \cite{Dragan2015EffectsCollaboration}, facial expressions \cite{Breazeal2005EffectsTeamwork}, lights \cite{Baraka2016EnhancingLights} and sounds \cite{Zahray2020RobotInteraction, Frid2022PerceptualGestures} have been proposed for HRI applications. In most prior work \cite{FernandezDeGorostizaLuengo2017SoundCues, Bellona2017EmpiricallyMovement, Roy2023TowardsSonification}, nonverbal communicative expressions are hand-crafted to convey a specific robot state, action, intention or emotion (SAIE).

When relevant to the context, nonverbal expressions can convey robot SAIEs with greater speed, universality, and appeal than spoken words \cite{Zhang2021BringingPerception, FernandezDeGorostizaLuengo2017SoundCues}. However, a trade-off exists between the complexity of the information conveyed and interpretability by diverse human collaborators, which may arise due to the subjective nature of these expressions. Miscommunications and decreased collaborative performance have been shown to occur even with seemingly intuitive nonverbal communication methods \cite{Fernandez2018PassiveInterpretability}.

In this work, rather than hand-crafting the communication strategy \cite{FernandezDeGorostizaLuengo2017SoundCues, Bellona2017EmpiricallyMovement, Roy2023TowardsSonification}, our aim is to learn how nonverbal communication should be structured to be best understood by users. The proposed approach uses nonverbal audio communication to convey functional robot states. A functional robot state describes the current task status of the robot (e.g. progressing, stuck, or accomplished). We auto-tune the acoustic parameter values (pitch bend, beats per minute, beats per loop) of nonverbal auditory expressions using a learning algorithm to convey three functional robot states (accomplished, progressing, stuck). An example real-world HRI scenario where nonverbal audio is advantageous is search and rescue. A robot must be able to rapidly convey functional states (e.g. danger, survivor found) to first responders. When visual modalities are limited due to smoke and debris, audio communication enables succinct HRI.

\begin{figure}[!t]
    % \vspace{-0.25cm}
    \centering
    % \hspace{-0.2cm}
    \includegraphics[width=1\linewidth,keepaspectratio]{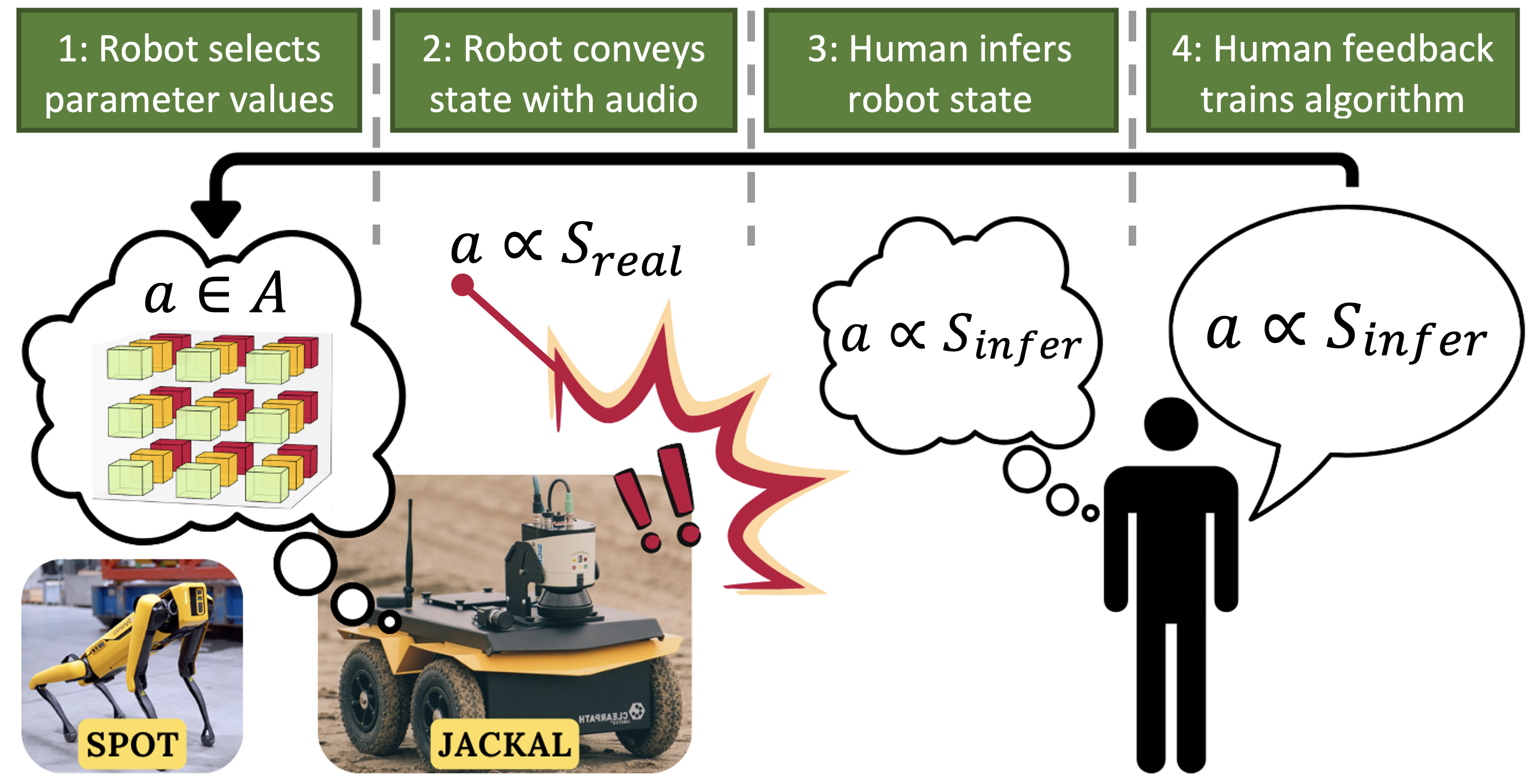}
    \caption{A robot nicknamed Jackal learning to generate nonverbal expressions to be correctly understood by users. The robot state perceived by the human is used to generate feedback to train the learning algorithm. The Spot robot image used within our user study procedure is also shown for reference.}
    \label{fig:communication_flow}
    \vspace{-0.6cm}
\end{figure}
 
This paper makes the following contributions: (i) We propose a learning-based framework for expressing functional robot states with nonverbal communication. We employ a reinforcement learning (RL) algorithm utilizing human feedback to customize expressions for individual users, aiming to enhance state classification accuracy. Through a user study, we demonstrate that this learning-based framework leads to significant improvements in users' ability to classify the state of the robot. (ii) We show that an algorithm initialization informed by previous user data can be used to speed up the learning process. (iii) We show that similarities exist between humans when interpreting robot states from nonverbal expressions. A high-level visualization of our approach is shown in Fig.\ref{fig:communication_flow} and discussed in detail in Section \ref{sec:approach}.

\section{Related Work}\label{sec:related_work}
Nonverbal communication is a valuable dimension of HRI, aiding in the formation of trust and social relations \cite{Breazeal2005EffectsTeamwork, Zinina2020Non-verbalLikeability}. In addition, the use of nonverbal communication in HRI has been shown to be an effective method for improving user experience and shared task performance \cite{Frid2022PerceptualGestures, Baraka2016EnhancingLights, Dragan2015EffectsCollaboration}.

\medskip\textbf{Nonverbal Auditory Expressions and Sonification:}
As applied to robotics, \textit{sonification} is the process by which sounds are used to represent robot states, actions, intentions and emotions (SAIEs). \textit{Sonification mapping} is the process by which SAIEs and sounds are related. \textit{Sonification techniques} such as juxtaposing rhythmic vs. continuous sounds \cite{Frid2022PerceptualGestures}, modelling nonverbal sounds \cite{FernandezDeGorostizaLuengo2017SoundCues} and using auditory sound emblems \cite{Zahray2020RobotInteraction} have been used to convey specified robot SAIEs. In many cases the mapping functions associating sounds to SAIEs are fixed, 1-to-1 mapping strategies which map a single SAIE to a single hand-tuned sound. In order to map three gestures to three emotions, Frid and Bresin manually applied filters and adjusted the acoustic parameters of rhythmic and continuous sounds \cite{Frid2022PerceptualGestures}. Comparably, Luengo et al. were able to effectively convey nine communicative expressions by hand-crafting nine detailed nonverbal sounds \cite{FernandezDeGorostizaLuengo2017SoundCues}. Citing a similar approach \cite{Bellona2017EmpiricallyMovement}, Zahray et al. manually mapped seven auditory sound emblems to seven of the Shimon robot's unique gestures \cite{Zahray2020RobotInteraction}.

Moving beyond 1-to-1 sonification mapping strategies, our previous work explored a parameterized state sonification model based on nonverbal sounds \cite{Roy2023TowardsSonification}. In this work, the acoustic parameters of a single sound were manually modulated to effectively express five functional robot states. We found that linearly modulating acoustic parameters of a sound is a sufficient strategy for communicating functional robot states; however, manually selecting and tuning parameters for each state is impractical and often sub-optimal. These results align with the limitations associated with manual sonification mapping strategies \cite{Frid2022PerceptualGestures, FernandezDeGorostizaLuengo2017SoundCues, Zahray2020RobotInteraction}; namely subjectivity, scalability and adaptability. Manually selected sounds might not universally convey an intended state, and may not scale to different robot morphologies. Building on the results of our previous work, this paper presents a learning-based approach that generalizes to expressing novel robot states without the need to hand-tune individual nonverbal sounds.

\medskip\textbf{Learning to Communicate from Human Feedback:}
Reinforcement learning (RL) \cite{SuttonRichardBarto2020ReinforcementLearning} has proven to be a useful method in HRI, enabling robots to learn interaction skills conducive to human-robot collaboration. Through RL-based methods, robots have learned social behaviours such as personalized proxemics \cite{Patompak2020LearningInteraction} and socially appropriate robo-waiter behaviours \cite{McQuillin2022LearningFeedback, Tseng2018ActiveFeedback}. Qureshi et al. demonstrated that a robot can use RL to learn basic human social interaction skills such as a hand-shake \cite{Qureshi2016RobotLearning}.

A recent survey on RL for social robots \cite{Akalin2021ReinforcementRobotics} finds that the majority of works focus on learning social behaviours while few approach the task of learning to communicate. Learning social behaviours provides robots with an understanding of human social norms and contextually appropriate responses such as maintaining personal space \cite{Patompak2020LearningInteraction, McQuillin2022LearningFeedback}. In contrast, learning to communicate involves the robot's ability to convey specific robot state-based information to humans. To effectively learn to communicate from human feedback, robots must convey state-based information legibly (i.e. in a clear and unambiguous manner)\cite{Dragan2015EffectsCollaboration, Dragan2013LegibilityMotion}. Improving the clarity of this information through legible expression helps humans accurately interpret the robot's state. Existing research exploring methods for learning to communicate primarily centers on facial expressions \cite{Churamani2018LearningModulations, Leite2012ModellingCompanions}, robot gaze \cite{Lathuiliere2019NeuralInteraction}, and often relies on pre-rendered expressions \cite{Papaioannou2017HybridLearning}. Much of the prior work focuses on learning a general \textit{one-size-fits-all} interaction strategy, without considering an individualized approach. In this work, we propose an RL-based approach to learn individualized communication strategies. To the best of our knowledge, we are the first to present an RL-based framework for generating personalized nonverbal auditory expressions to convey a robot's internal state.

\section{Generating Auditory Expressions using RL}\label{sec:approach}
In our prior research, we introduced a nonverbal communication model involving manual modulation of acoustic parameters to convey functional robot states \cite{Roy2023TowardsSonification}. In this paper, we propose a learning-based approach which automatically learns these parameters to produce nonverbal expressions for communicating functional robot states.

Following the structure of our previously developed parameterized communication model \cite{Roy2023TowardsSonification}, we created a sound library by modulating a fixed number of acoustic parameters ($P$) of an audio sample set on a loop. Each acoustic parameter was discretized into a unique number of regions $D_i$, resulting in a sound library of size: $\prod_{i=1}^{P} D_i$. This sound library was framed as the action space for a multi-armed bandit (MAB) RL algorithm \cite{SuttonRichardBarto2020ReinforcementLearning, Slivkins2019IntroductionBandits}, in which each sound represents a possible action the RL agent can execute. This MAB algorithm comprises an RL agent (robot), its action space (all possible actions $a \in A$ which the agent can take) and a reward signal $R$ based on feedback from the user.

In this work, the robot explores combinations of acoustic parameter values to generate personalized sounds which elicit the highest reward signal from individual humans. The goal of the robot in state $S_{real}$ is to generate a nonverbal sound which elicits a congruence between the human's perceived state ($S_{infer}$) and the true state of the robot ($S_{infer}=S_{real}$). This approach is visualized in Fig.\ref{fig:communication_flow}.

\subsection{Proposed Framework}\label{subsec:proposed_framework}
We propose a reinforcement learning (RL)-based approach that exploits noisy human feedback to auto-tune acoustic parameter values correlated to an intended robot state. This framework iteratively presents users with a nonverbal audio expression signifying a given state ($S_{real}$), collects user feedback specifying the state they infer the robot to be in ($S_{infer}$), uses this user feedback to formulate a reward signal ($R$), and integrates this reward signal into an MAB algorithm known as UCB1 (Upper Confidence Bounds algorithm \cite{Kuleshov2014AlgorithmsProblems, Lattimore2020BanditAlgorithms}) to update the acoustic parameters of the audio expression signifying $S_{real}$. UCB1 is a confidence-based algorithm that balances the exploitation of a known immediate reward with exploration to maximize long-term cumulative gain. Using UCB1, we posit that values of each acoustic parameter ($P$) can be iteratively tuned to produce a nonverbal sound which is accurately understood by the user ($S_{infer}=S_{real}$). This algorithm pseudo-code is shown in Algorithm \ref{ucb1_pseudocode}.

We formulate this problem as an MAB problem firstly due to its lack of sequenced actions ($a_1, a_2, ... a_n$), which is characteristic of bandit algorithms. Secondly, the reward signal for each action is an unknown distribution as it derives from noisy human feedback, which justifies the selection of a confidence-based MAB algorithm: UCB1 \cite{Kuleshov2014AlgorithmsProblems, Lattimore2020BanditAlgorithms}. Users may be presented with the same sound twice and may not necessarily input identical responses, resulting in a varying reward signal ($R$). This reward uncertainty results in an unknown reward distribution, as opposed to a known fixed reward used in typical Q-learning implementations \cite{SuttonRichardBarto2020ReinforcementLearning}.

To instantiate Algorithm \ref{ucb1_pseudocode}, we initialize a Q-value table of size $\prod_{i=1}^{P} D_i$ associating a value ($Q$) to each action ($a$) within the action space ($a \in A$). The value of each action, also known as the \textit{Q-value}, signifies the estimated reward ($R$) that would be observed if that action was executed ($a_t = a$). If a system is learning to communicate multiple robot states, a unique Q-table will be initialized for each state. This algorithm iteratively selects an action by taking an \textit{argmax} of all possible Q-values $Q_t(a)$ summed with their respective uncertainty terms $U_t(a)$. If multiple values are equally maximal (e.g. at initialization), one max-value action is selected at random.

After selecting an action (sound presented to the user), the user is prompted to indicate their inferred state of the robot ($S_{infer}$) and associated confidence ($C$) in their response. This feedback is used to formulate a scaled reward signal ($R$) based on \cite{Wilde2021LearningFeedback}'s scaling technique. The formulation of this reward signal is outlined in line 12 of Algorithm \ref{ucb1_pseudocode}. This reward signal is used to update the executed action's Q-value $Q_{t+1}(a)$ (line 13 of Algorithm \ref{ucb1_pseudocode}). After updating $Q_{t+1}(a)$, the algorithm continues to explore the action space until it reaches convergence or exhausts its available budget $B$.

\begin{figure}[!b]
    \vspace{-0.85cm}
    \centering
    % \hspace{-0.2cm}
    \includegraphics[width=1.0\linewidth,keepaspectratio]{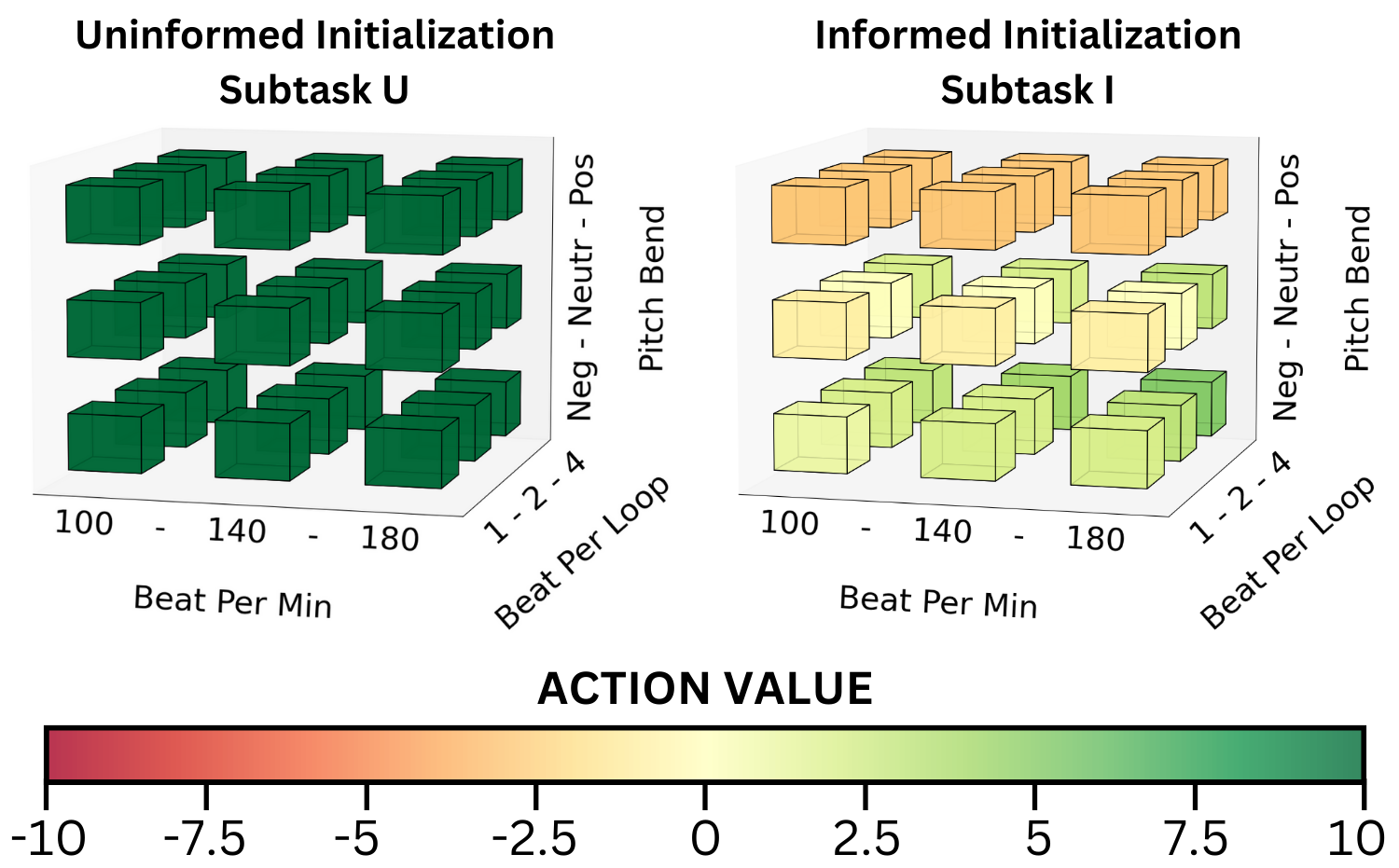}
    \caption{Two Q-table initializations for a sound library with $P=3$ parameters, each discretized into $D=3$ regions. The resulting Q-table has 27 actions, each represented as a coloured cube denoting that action's estimated reward. (Left - Subtask U - Uninformed Initialization) Each Q-value is initialized to its maximum value following the principle of optimism under uncertainty. In this uninformed example, all values are initialized to 10. (Right - Subtask I - Informed Initialization) Q-values are set to predefined values derived from prior knowledge. In this informed example, sounds with a negative pitch bend and a greater number of beats per loop are initialized with a positive value.}
    \label{fig:matrix_init_u_i}
    % \vspace{-0.3cm}
\end{figure}

% LATEX PSEUDO_CODE
\begin{algorithm}[!b]
    \footnotesize % \small, \footnotesize, \scriptsize, \tiny
    \setstretch{1.05}
    \caption{UCB1 Algorithm based on Human Feedback\smallskip}\label{ucb1_pseudocode}
        \begin{algorithmic}[1]
            \State \textbf{initialize} $N_t(a)=0$ and $Q_t(a)$ for all values in Q-table
            \State \textbf{initialize} $t=0$, $F=0$
            \State \textbf{set} hyper-parameters $z$, $B$, $F_{conv}$, $\Delta Q_{conv}$
            \For{iterations in budget $B$}
                \State \textbf{increment} time step $t \gets t-1$
                \State \textbf{calc} each action's uncertainty $U_t(a) = z\sqrt{\frac{2log(t)}{N_t(a)}}$
                \State \textbf{select} action $a_t = \underset{a \in A}{\argmax}(Q_t(a) + U_t(a))$
                \State \textbf{increment} $N_t(a) \gets N_t(a)-1$
                \State \textbf{execute} action $a_t$
                \State \textbf{probe} user for feedback $S_{infer}$ and $C$
            
                \State \textbf{calc} $S_{check} = 
                    \begin{cases}
                    1& \text{\small if correct $S_{infer}=S_{real}$} \\
                    -1& \text{\small if incorrect $S_{infer} \neq S_{real}$} \\
                    \end{cases}
                    $\;
                \State \textbf{calc} reward $R = S_{check} \cdot C$
                \State \textbf{update} $Q_{t+1}(a) = \left[ \left( 1-\frac{1}{N_t(a)}\right) \cdot Q_{t}(a) \right] + \left[ \frac{1}{N_t(a)} \cdot R \right]$
            
                \State \textbf{calc} $F =
                    \begin{cases}
                    F+1& \text{\small if $(S_{infer} = S_{real}) \oplus (a_t = a_{t-1})$} \\
                    F+1& \text{\small if $(S_{infer} = S_{real}) \oplus (\Delta Q \le \Delta Q_{conv})$} \\
                    0& \text{\small otherwise} \\
                    \end{cases}
                    $\;
                \State 
                \If {$F \leq F_{conv}$} 
                    \State \textbf{break}
                \EndIf
            \EndFor
            \State \textbf{terminate}
            \algrule             % \\\hrulefill
            \Statex \textbf{VARIABLE LEGEND}
            \Statex \text{$t =$ timestep}
            \Statex \text{$a_t =$ selected action $a$ at time $t$}
            \Statex \text{$Q_{t}(a) =$ estimated value of action $a$ at time $t$}
            \Statex \text{$U_{t}(a) =$ uncertainty of action $a$ at time $t$}
            \Statex \text{$N_{t}(a) =$ number of times $a$ has been selected}
            \Statex \text{$R =$ reward generated by $S_{check}$, $S_{infer}$, $C$}
            \Statex \text{$S_{check} =$ $\pm1$ based on $S_{infer}$ and $S_{real}$}
            \Statex \text{$S_{infer} =$ state inferred by the user}
            \Statex \text{$S_{real} =$ real state of the robot}
            \Statex \text{$C =$ user's confidence in response $\in [0, 10]$}
            \Statex \text{$F =$ convergence counter}
            \Statex \text{$\Delta Q =$ difference between $Q_{t}(a)$ and $Q_{t+1}(a)$}
            \algrule             % \\\hrulefill
            \Statex \textbf{HYPER-PARAMETERS LEGEND}
            \Statex \text{$z = 0.5 =$ exploration vs exploitation factor}
            \Statex \text{$B = 60 =$ max iterations budget}
            \Statex \text{$F_{conv} = 3 =$ $F$ convergence threshold}
            \Statex \text{$\Delta Q_{conv} = 2.0 =$ threshold to increment $F$}
    \end{algorithmic}
\end{algorithm}

Q-table initialization plays a large role in how the action space is explored. We investigate two methods for Q-table initialization in this work: uninformed initialization and informed initialization. Fig.\ref{fig:matrix_init_u_i} visualizes both Q-table initialization styles. Following an uninformed initialization, all values within the Q-table are initialized to their maximum possible value (in this case $10$) following the principle of \textit{optimism under uncertainty} \cite{Slivkins2019IntroductionBandits}. With all Q-values initialized to their maximum, the algorithm's $argmax$ function (line 7 of Algorithm \ref{ucb1_pseudocode}) forces the exploration of all actions before returning to those yielding the highest rewards. To streamline Q-table updates, we propagate the reward signal from a single action exploration to update corresponding Q-values across all robot states in the learning process. When a user associates a sound with the parameter combination $[1, 2, 0]$ to denote the "Stuck" state, the corresponding Q-value in the "Stuck" Q-table is updated positively, whereas Q-tables for all other states receive negative updates.

Under informed initialization, the Q-values are set to predefined values derived from prior knowledge. These predefined values range from -10 to 10 and were derived from user-suggested sound representations for each robot state in prior research \cite{Roy2023TowardsSonification}. In informed initialization, a green cube (positive value) signifies users' previous identification of a sound as appropriate for the target state, whereas an orange cube (negative value) indicates a sound was deemed less appropriate. This method utilizes the algorithm's $argmax$ function (line 7, Algorithm \ref{ucb1_pseudocode}) to prioritize user-suggested sounds, initially presenting those with high initial values. Informed initialization has the potential to reduce the learning steps required for convergence, assuming alignment between the new user's preferences and those of pilot users.

We employ a convergence approach based on a convergence counter ($F$). Convergence is achieved when this counter ($F$) surpasses the convergence threshold ($F_{conv}$). The convergence counter ($F$) is incremented each iteration based on conditions specified in line 14 of Algorithm \ref{ucb1_pseudocode}. $F$ returns to zero when neither of these conditions is met, ensuring convergence only occurs when the best action in the Q-table is selected and correctly classified three times consecutively. We empirically choose a  threshold ($F_{conv}=3$) as a tradeoff between shorter experiment time and convergence stability. When a system learns to communicate multiple robot states, each state has its own convergence counter $F$.

\section{User Study}\label{sec:user_study}
To validate the developed approach we conducted a screen-based HRI study. This user study was reviewed and approved by the Monash University Human Research Ethics Committee (MUHREC) with project ID 37157.

\begin{figure}[!b]
    \vspace{-0.5cm}
    \centering
    % \hspace{-0.2cm}
    \includegraphics[width=1\linewidth,keepaspectratio]{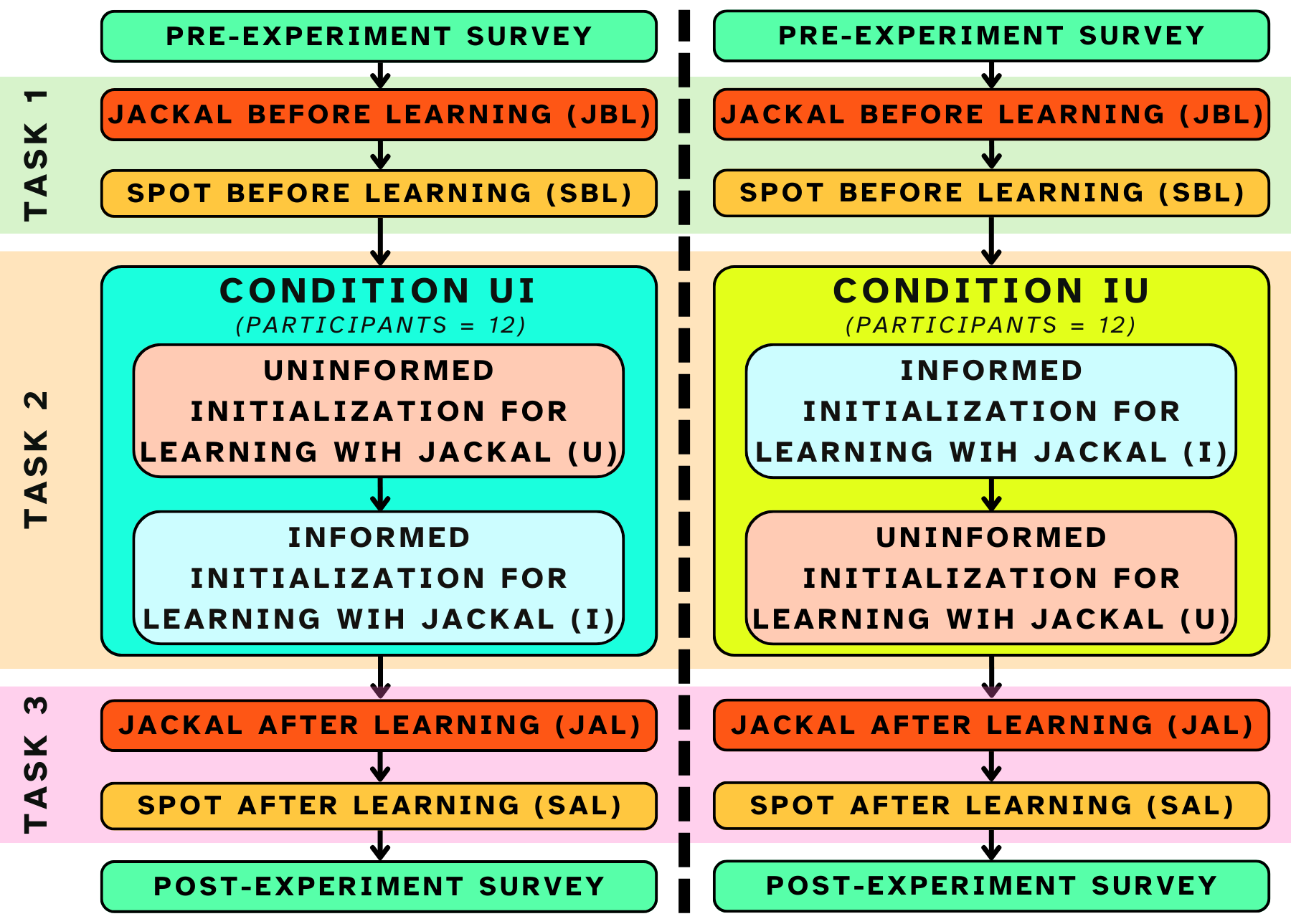}
    \caption{Flowchart depicting the tasks and conditions of the user study.}
    \label{fig:study_plan_flowchart}
    % \vspace{-0.5cm}
\end{figure}

\subsection{Procedure}\label{subsec:procedure}
The study procedure encompassed three components: a pre-experiment demographic survey, an interactive screen-based survey where users interacted with two robots nicknamed Jackal and Spot, and a post-experiment feedback survey. Fig.\ref{fig:study_plan_flowchart} shows a flow chart of the study design. The interactive survey comprised three tasks, each divided into two subtasks: (1) Jackal Before Learning and Spot Before Learning, (2) Uninformed Initialization and Informed Initialization, (3) Jackal After Learning and Spot After Learning. To fully define our experimental setup, we provide a link to our public codebase\footnote{Codebase: \url{https://github.com/liamreneroy/RL_audio}}. Within this codebase, we provide the original interactive survey (Jupyter Notebook) with added step-by-step markdown comments describing the study procedure to enable reproducibility. Following this procedure, Task 1 evaluated participants' proficiency in discerning functional states from nonverbal expressions before running the algorithm. In Task 2 the algorithm attempted to learn suitable acoustic parameters tailored to individual participants. Task 3 served as a post-algorithm assessment to compare users' accuracy in correctly inferring states after running the algorithm.

In Task 1, users were presented with an image of two robots; a rover robot nicknamed Jackal and a quadruped robot nicknamed Spot. Both images of these robots are viewable in Fig.\ref{fig:communication_flow}. Task 1 involved two subtasks: Jackal Before Learning (JBL) and Spot Before Learning (SBL). In both subtasks, the robot expressed each of its three functional states (\textit{stuck}, \textit{accomplished}, \textit{progressing}) using the communication model developed in the previous user study \cite{Roy2023TowardsSonification}. Two similar audio samples (short duration, consistent pitch and volume) sourced from the Ableton Live 10 sample library were used to generate Jackal and Spot's parameterized sound libraries. Original audio samples and full sound libraries are available in supplementary material and the linked codebase. Our methodology for modulating the acoustic parameters of a base sound to produce a sound library is outlined in our prior work \cite{Roy2023TowardsSonification}.

In this study, each sound library was generated by linearly modulating three ($P=3$) widely studied acoustic parameters \cite{FernandezDeGorostizaLuengo2017SoundCues} of a base sound set on a loop: beats per minute (BPM), beats per loop (BPL) and pitch bend. The BPM parameter adjusted the speed at which the audio loop was played. The BPL parameter adjusted the number of times the base sound was added to the loop. The pitch bend parameter modulated the pitch of the audio loop with a positive or negative inflection. Each parameter was discretized into 3 regions ($D=3$), resulting in a 3-dimensional sound library with 27 ($\prod_{i=1}^{P} D_i = 3\cdot3\cdot3 = 27$) unique sounds. By following an identical acoustic parameterization structure using different base sounds, both robots communicated with the same language structure and each robot had its own distinct "voice". Users commented that the Jackal's sounds (library A) resembled musical beeps, while the Spots's resembled retro video game beeps. We chose different sounds for the two robots to avoid confusion and create a distinction between the Jackal and Spot when presenting sounds to the users. A visualization of this sound library is shown in Fig.\ref{fig:matrix_init_u_i} where each sound is represented as a coloured cube.

The initial parameter values selected for each state were derived from user-suggested sound representations for each robot state in prior research \cite{Roy2023TowardsSonification}. The state \textit{stuck} was expressed using a high number of beats per loop and a negative pitch bend. The state \textit{accomplished} was expressed using a moderate number of beats per loop and a positive pitch bend. The state \textit{progressing} was expressed using a low number of beats per loop and no pitch bend. The BPM was initialized as neutral (140BPM in Fig.\ref{fig:matrix_init_u_i}) for all states as this parameter was not used in our previous user study. After each expression, users were prompted to indicate their inferred state of the robot and associated confidence in their response. Users were allowed to replay the sounds as many times as they wanted, with no time limit.

Following Task 1, participants were randomly assigned Subtask Uninformed Initialization (U) or Informed Initialization (I). This randomization aimed to mitigate potential ordering bias. Participants who performed Subtasks U then Subtask I followed Condition UI, while those who completed Subtasks I then Subtask U followed Condition IU. In both subtasks, the Jackal used Algorithm \ref{ucb1_pseudocode} to learn suitable parameter values for three functional states. Subtasks U and I differed in how the Q-tables for each robot state were initialized, as described in Section \ref{subsec:proposed_framework} and visualized in Fig.\ref{fig:matrix_init_u_i}. The algorithm in Subtasks U/I iteratively updated each state's Q-table until one of two conditions was met: the system reached convergence ($F = F_{conv}$) or used up its budget $B$. As denoted in line 4 of Algorithm \ref{ucb1_pseudocode}, the algorithm terminates if a participant exhausts their allotted budget. In this case, the final parameter combination assigned to each robot state is that with the highest action value in each respective Q-table during the final learning iteration (iteration 60 since $B=60$). Algorithm \ref{ucb1_pseudocode} lists hyper-parameters, their values, and their role in algorithm convergence, which were selected empirically and fixed for all participants. The supplementary video accompanying this paper illustrates the algorithm's convergence from the state "stuck" using an uninformed initialization (Subtask U).

The final task, consisting of Subtasks Jackal After Learning (JAL) and Spot After Learning (SAL), was near-identical to Task 1. Users were again presented with images of two robots: Jackal (Subtask JAL) and Spot (Subtask SAL). In both subtasks, the robot communicated its three functional states (\textit{stuck}, \textit{accomplished}, \textit{progressing}) using parameters learned from the algorithm in Task 2. If a participant completed condition UI, the parameters learned in Subtask I were used for Task 3. If they completed condition IU, then the parameters learned in Subtask U were used for Task 3. The results of Task 3 were used to compare a user's accuracy in correctly inferring functional states based on nonverbal expressions before and after running the algorithm.

\subsection{Participants}\label{subsec:participants}
This study comprised 24 participants, drawn from Monash University students and staff on campus, as well as external non-affiliated individuals. Given the unknown effect size, we did not conduct a power analysis. However, drawing from a comparable study in which a robot learned socially appropriate behaviours through real-time user feedback with statistically significant results (21 participants) \cite{McQuillin2022LearningFeedback}, we estimated that 24 would be a suitable number of participants for our research. Demographic analysis showed 68\% engaged in daily music or instrument activities, and 80\% had experience in robot-related communication. 84\% of participants were 18-29 years old, and 16\% were 30-49 years old.

\subsection{Hypotheses}\label{subsec:hypotheses}
To achieve our aim of learning how nonverbal communication should be structured to be best understood by users, we formulate and evaluate three hypotheses. Using the proposed learning algorithm, we hypothesized that the robot would learn to communicate functional states, allowing a human to more accurately infer the robot's functional state (\textbf{H1}). This was tested by comparing the degree of state classification accuracy in Task 3 compared to Task 1. Our second hypothesis (\textbf{H2}) was that an informed initialization would reduce the required learning steps for algorithm convergence. This was tested by comparing the number of learning steps users took to reach convergence between Subtask U (Uninformed Initialization) and Subtask I (Informed Initialization). Our final hypothesis (\textbf{H3}) was that there would be similarities between the models learned for different humans. We tested this hypothesis by analyzing participants' final parameter values across all three states.

\section{Results}\label{sec:results}

\subsection{Improved Classification Accuracy (H1)}\label{subsec:algorithm_conv}
To test whether the proposed algorithm could learn to communicate accurately interpreted nonverbal auditory expressions using noisy human feedback (\textbf{H1}), we compared users' accuracy in correctly inferring functional states before and after running the learning algorithm.  We employed logistic regression to analyze pre- and post-learning data. The logistic regression model is represented by Equation \ref{eq:prior_post_logistic_regression}, where the predicted variable \textit{Correct} signifies correct ($\text{Correct}=1$) or incorrect ($\text{Correct}=0$) state identification by users. All variables in the model were categorical, and a test for interaction between variables revealed no significant interactions. Table \ref{tab:prior_post_regression_table} summarizes the results of this regression.
\vspace{-0.15cm}
\begin{equation}
    \resizebox{0.44\textwidth}{!}{$
    \text{Correct} = \beta _0 + \beta _1\text{State} + \beta _2\text{LearnStage} + \beta _3\text{TrainCond} + \beta _4\text{Robot}
    $}
    \label{eq:prior_post_logistic_regression}
\end{equation}

\begin{figure}[!b]
    \vspace{-0.5cm}
    \centering
    % \hspace{-0.2cm}
    \includegraphics[width=1\linewidth,keepaspectratio]{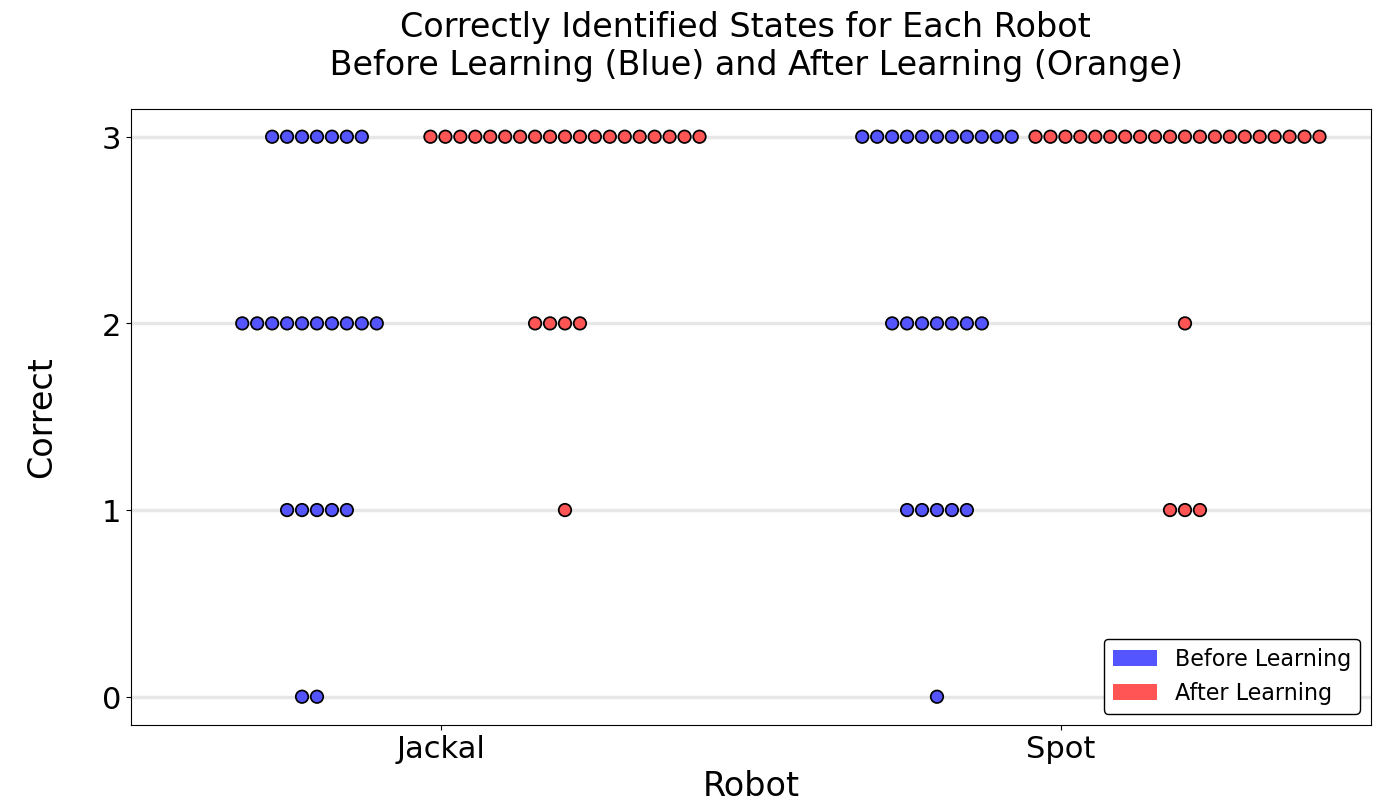}
    \caption{Dot plot depicting the number of states users were able to identify correctly before learning (blue) and after learning (red) with two different robots: Jackal and Spot.}
    \label{fig:correct_states_violin_plot}
\end{figure}

Analyzing the results shown in Table \ref{tab:prior_post_regression_table}, users demonstrated a significant improvement in state recognition accuracy after learning. Users' state recognition accuracy improved for both the Jackal (used during learning) and Spot (not used during learning). This improvement is visualized in Fig.\ref{fig:correct_states_violin_plot}. Furthermore, the users' ability to accurately discern the robot's state was significantly influenced by its state, suggesting that certain robot states were easier to distinguish than others.

% \vspace{-0.3cm}

\begin{table}[!b]
    \vspace{-0.25cm}
    \centering
    \caption{Results of logistic regression model to analyze the variables which contribute to users' accuracy in correctly inferring functional states.}
    \resizebox{\columnwidth}{!}{%

    \begin{tabular}{|l|c|c|c|}
    \hline
    \textbf{Variable} & \textbf{Coefficient Est.} & \textbf{p-Value} & \textbf{Significance}\\ \hline
    
    % (Intercept) & -1.9433 & 1.61e-06 & \textbf{***}\\ 
    % \hline 
    State=Progres. & -0.3147 & 0.37664 & \\ 
    \hline 
    State=Stuck & -1.0418 & 0.00949 & \textbf{**}\\ 
    \hline 
    LearnStage=Before & 1.5983 & 3.98e-06 & \textbf{***}\\ 
    \hline 
    TrainCond=UI & 0.2401 & 0.43949 & \\ 
    \hline 
    Robot=Spot & -0.2401 & 0.43949 & \\ 
    \hline 
    \multicolumn{4}{|c|}{Significance Codes: \quad 0 $<$ \textbf{***} $<$ 0.001 $<$ \textbf{**} $<$ 0.01 $<$ \textbf{*} $<$ 0.05 }\\
    \hline 
    \end{tabular}
    }
    \label{tab:prior_post_regression_table}
\end{table}
% \vspace{-0.3cm}

\begin{table*}[!b]
    \vspace{-0.3cm}
    \centering
    \caption{Results of two Wilcoxon Signed Rank Tests. The ranked sum signifies the number of times that any value in the compared dataset is larger than a value in the reference dataset, where ties (e.g. 2 vs. 2) are broken by adding 0.5 to the ranked sum. The theoretical max ranked sum is $12*12=144)$ as each data series contains 12 data points (one for each participant in either condition UI or IU). Tests 1 and 2 both yield statistical significance ($P<0.05$).}
    \resizebox{\textwidth}{!}{%

    \begin{tabular}{|c|c|c|c|c|c|c|c|}
    \hline
    \textbf{Test}&\textbf{Reference Dataset}&\textbf{Mean}&\textbf{Compared Dataset}&\textbf{Mean}&\textbf{Shift}&\textbf{Ranked Sum}&\textbf{p-Value}\\\hline   
    1 & Informed Init Steps CondUI & 15.58 & Uninformed Init Steps CondUI & 39.08 & 23.50 & 137 / 144 & 0.00018 \\ \hline

    2 & Informed Init Steps CondIU & 17.92 & Uninformed Init Steps CondIU & 41.25 & 23.33 & 134 / 144 & 0.00032 \\ \hline 
    \end{tabular}
    }
    \label{tab:wilcoxon_tests_table}
\end{table*}

This result supports hypothesis \textbf{H1}, as users were able to more accurately identify states when the robot's communication strategy was tailored specifically to their individual preferences. A key observation is that a similar improvement in classification accuracy is observed with the Jackal (Subtask JBL to JAL) and Spot (Subtask SBL to SAL), despite users only training with Jackal sounds. The Jackal and Spot sound libraries were developed using the same parameterization structure but two distinct base audio samples, therefore, users' classification accuracy may have differed when the parameters learned using the Jackal sound library were ported to the Spot sound library. This similar improvement in classification accuracy with both robots suggests that learned acoustic parameters can generalize to alternate sounds which users have not explored in-depth and still produce similar improvements in state classification accuracy.

\subsection{Decreased Learning Steps (H2)}\label{subsec:rate_of_conv}
\begin{figure}[!b]
    \vspace{-0.45cm}
    \centering
    % \hspace{-0.2cm}
    \includegraphics[width=1\linewidth,keepaspectratio]{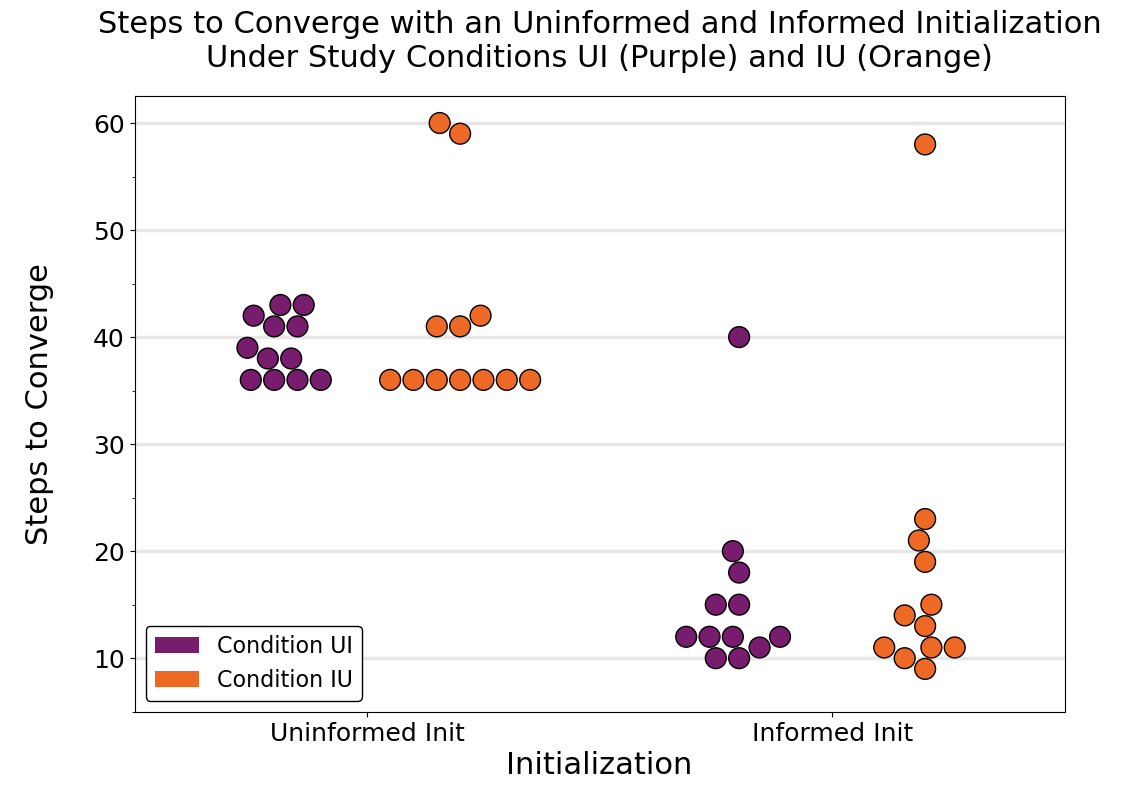}
    \caption{Dot plot depicting the number of steps the learning algorithm took to converge for each user from Subtask U (Uninformed Init) and Subtask I (Informed Init) under study conditions UI (Purple) and IU (Orange).}
    \label{fig:congerge_steps_plot}
\end{figure}

To test whether an informed initialization could be used to speed up the learning process (\textbf{H2}), we compared the number of learning steps taken to reach convergence between Subtask U (Uninformed Initialization) and Subtask I (Informed Initialization). These data are visualized in Fig.\ref{fig:congerge_steps_plot}. Two non-parametric Wilcoxon Signed Rank tests were performed to compare the results of both subtasks under each study condition (UI and IU - see Fig.\ref{fig:study_plan_flowchart}). The Wilcoxon Signed Rank Test was chosen for data analysis due to non-normal data distribution, within-subject data collection, and the examination of two data series with a single independent variable. Test results are summarized in Table \ref{tab:wilcoxon_tests_table}.

The informed initialization (Subtask I) resulted in significantly faster convergence of the algorithm compared to an uninformed initialization (Subtask U). The average number of steps necessary to converge with an uninformed initialization (mean=39.08) decreased by 23.5 (60.1\%) when an informed initialization was implemented (mean=15.58). This result supports our hypothesis \textbf{H2}. Although there is no guarantee that an algorithm initialization generated from previous user data will align with new users, the results of Task 2 demonstrate that an informed initialization results in a reduced number of steps to convergence, while retaining recognition performance accuracy, as seen in Table \ref{tab:wilcoxon_tests_table}.

\subsection{Similar Converged Parameter and Suggestibility (H3)}\label{subsec:suggestibility}
To investigate if similarities exist between humans when interpreting robot states from nonverbal expression (\textbf{H3)}, we analyzed participants' final parameter values across all three states. These data were used to generate 3D heatmaps shown in Fig.\ref{fig:parameter_heatmaps}. Visualizing these heatmaps revealed that the method used for algorithm initialization (informed vs. uninformed) strongly influenced the degree to which users converged on similar final parameter values. In Fig.\ref{fig:parameter_heatmaps}, right-hand plots show that users can be guided to a similar final parameter configuration using informed initialization, while the left-hand plots display greater diversity in final parameter values in the absence of an informed initialization.

\subsection{Influence of Model Parameters}\label{subsec:additional_results}
\begin{figure}[!t]
    % \vspace{-0.45cm}
    \centering
    % \hspace{-0.2cm}
    \includegraphics[width=1\linewidth,keepaspectratio]{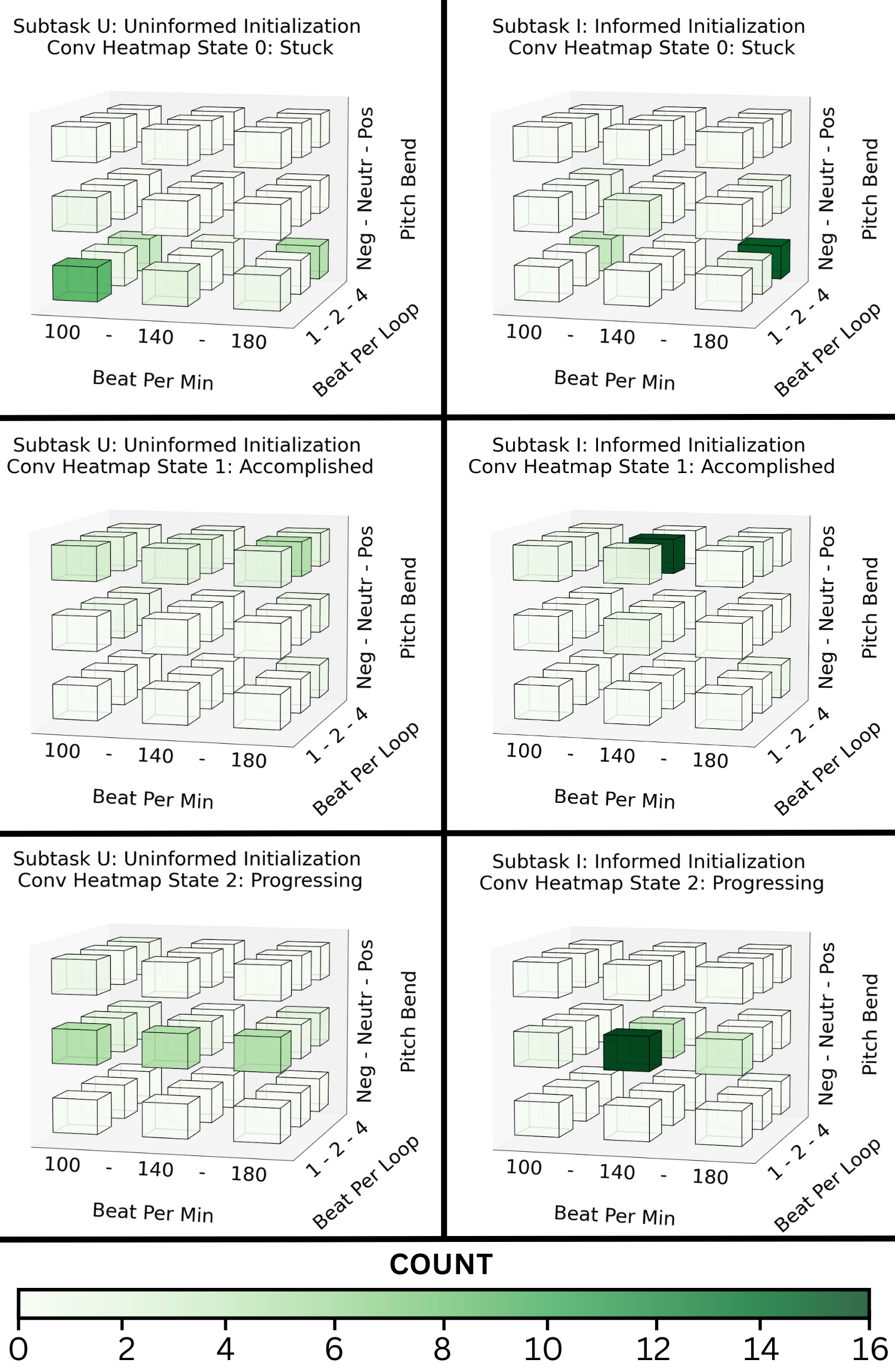}
    \caption{Heatmaps depicting the number of users who converged on a combination of acoustic parameter values for all three robot states. The left-hand plots depict three states starting from an uninformed initialization, while the right-hand plots depict three states starting from an informed initialization. A darker cube represents a greater number of users who converged to that specific combination of acoustic parameter values.}
    \label{fig:parameter_heatmaps}
    \vspace{-0.75cm}
\end{figure}

A visual analysis of the heatmap plots in Fig.\ref{fig:parameter_heatmaps} revealed that participants favoured negative pitch bend for the state \textit{Stuck}, positive pitch bend for the state \textit{Accomplished}, and neutral pitch for the state \textit{Progressing}. Less clear distinctions were observed for beats per minute (BPM) and beats per loop (BPL). This suggests that pitch bend exhibited a stronger influence than BPM and BPL, revealing non-uniform parameter effects. To quantitatively validate this observation, we performed logistic regression analysis on each user's converged Q-table for each robot state. The regression model is expressed as Equation \ref{eq:influence_logistic_regression}, where the predicted variable \textit{PositiveVal} signifies a parameter combination having a positive ($\text{PositiveVal}=1$) or negative ($\text{PositiveVal}=0$) Q-value upon algorithm convergence. For instance, a positive Q-value at convergence ($\text{PositiveVal}=1$) in the \textit{Stuck} Q-table for the parameter combination [BPM, BPL, Pitch]=[1, 2, 0] indicates that the user inferred this sound to signify \textit{Stuck}. Table \ref{tab:influence_regression_table} summarizes the results of this regression.
\vspace{-0.1cm}
\begin{equation}
    \resizebox{0.44\textwidth}{!}{$
    \text{PositiveVal} = \beta _0 + \beta _1\text{BPM} + \beta _2\text{BPL} + \beta _3\text{PitchBend} + \beta _4\text{State} + \beta _5\text{Init}
    $}
    \label{eq:influence_logistic_regression}
\end{equation}

The results of this regression shown in Table \ref{tab:influence_regression_table} indicate that Pitch Bend is highly significant, aligning with the results visualized in Fig.\ref{fig:parameter_heatmaps}. In addition, BPL shows significance, while the BPM yields no significance. This numerical result indicates that parameters have varying influence on the users' perception of the robot's state. The regression also reveals that the initialization method (Uninformed vs. Informed) significantly influences the likelihood that a combination of parameters will achieve a positive Q-value at convergence. This result aligns with expectations due to the distinct settings distribution between the two initialization approaches.

\section{Discussion}\label{sec:discussion}
In line with existing literature on learning to communicate facial expressions \cite{Churamani2018LearningModulations, Leite2012ModellingCompanions} and robot gaze \cite{Lathuiliere2019NeuralInteraction}, our findings indicate that individualized nonverbal auditory expressions can be learned through human feedback. Prior work has found that a single design for an interactive system is often only suitable for a small subset of users \cite{Gajos2017}. As supported in this work, individualized systems have been shown to improve overall task performance and user satisfaction in human-robot/computer interaction scenarios \cite{Sekmen2013AssessmentInteractions, Gajos2010AutomaticallySupple}.

Our post-experiment survey responses highlight that a balance exists between effort and personalization when designing adaptive communication. While many users expressed excitement over the final learned sounds meeting their expectations, several found the learning process to be long, tedious, and excessively repetitive. An informed initialization can reduce the length of this learning process for new users.

The proposed approach utilizes a tabular RL method, which helps speed up convergence during online learning, but limits the ability to expand the action space; either by increasing the number of acoustic parameters $P$ or their degree of discretization $D$. In this study, the action space contained 27 actions ($\prod_{i=1}^{P} D_i = 3\cdot3\cdot3 = 27$), therefore, exploring each action one-by-one was viable. This approach would become cumbersome as the number of parameters $P$ or the degree of parameter discretization $D$ increased. 

\begin{table}[!t]
    \vspace{0.15cm}
    \centering
    \caption{Results of logistic regression model to analyze the variables which contribute to a combination of parameters (e.g. [BPM, BPL, Pitch]=[1, 2, 0]) having a positive Q-value when the algorithm converges.}
    \resizebox{\columnwidth}{!}{%

    \begin{tabular}{|l|c|c|c|}
    \hline
    \textbf{Variable} & \textbf{Coefficient Est.} & \textbf{p-Value} & \textbf{Significance}\\ \hline
    
    % (Intercept) & -0.7132 & 3.71e-11 & \textbf{***}\\ 
    % \hline 
    BPM=1 & -0.0173 & 0.8355 & \\ 
    \hline 
    BPM=2 & 0.0034 & 0.9669 & \\ 
    \hline 
    BPL=1 & 0.1875 & 0.0246 & \textbf{*}\\ 
    \hline 
    BPL=2 & 0.1739 & 0.0372 & \textbf{*}\\ 
    \hline 
    PitchBend=1 & 0.4498 & 4.71e-08 & \textbf{***}\\ 
    \hline 
    PitchBend=2 & -0.0108 & 0.8989 & \\ 
    \hline 
    State=Progres. & -0.0419 & 0.6161 & \\ 
    \hline 
    State=Stuck & 0.1195 & 0.1483 & \\ 
    \hline 
    Init=Uninformed & -0.3799 & 2.22e-08 & \textbf{***}\\ 
    \hline 
    \multicolumn{4}{|c|}{Significance Codes: \quad 0 $<$ \textbf{***} $<$ 0.001 $<$ \textbf{**} $<$ 0.01 $<$ \textbf{*} $<$ 0.05 }\\
    \hline 
    \end{tabular}
    }
    \label{tab:influence_regression_table}
\vspace{-0.15cm}
\end{table}

The finding of non-uniform parameter influence has the potential to be leveraged in real-time learning scenarios to address this known limitation. As high-influence parameters are identified, these parameters can be prioritized throughout action space exploration. An additional approach to decrease the time the user needs to spend training the system is to integrate strategies for generating rewards from task-focused human actions \cite{Ghadirzadeh2021Human-CenteredLearning, Qureshi2016RobotLearning, Weber2018HowLearning}.

Our findings suggest that commencing with a universal communication strategy and subsequently tailoring it to individual needs can elevate performance. To simplify selecting an effective starting point, future research could explore data-driven methods to identify high-influence patterns across users, thereby refining initialization strategies. Our results, consistent with previous findings \cite{Roy2023TowardsSonification}, also show that some functional states are inherently difficult to convey through nonverbal audio alone. For clearer communication of task status, integrating audio with other nonverbal modalities (e.g. motion \cite{Venture2019RobotMotions}), may enhance understanding. For audio-only systems, generating a diverse sound library may enhance the system's ability to convey different states.

\textbf{Limitations}: As mentioned in Section \ref{subsec:proposed_framework}, we select a convergence threshold ($F_{conv}=3$) empirically. We observe that only 1 of 24 participants did not converge by the end of the allotted 60-step budget. Nevertheless, this criterion should be further optimized in future works. Our participant pool was not representative of the average population, predominantly comprising young, educated individuals with prior experience in robot-related communication. Further research is needed to examine if these findings generalize to more diverse populations. Finally, we did not explore the relationship between a robot's embodiment and the sounds it produced. Future research should examine how a robot's physical appearance affects perceptions of its nonverbal sounds and the resulting learning outcomes.

\section{Conclusion and Future Work}\label{sec:conclusion}
In this work, we proposed a reinforcement learning (RL) algorithm capable of using noisy human feedback to learn to communicate three functional robot states (accomplished, progressing, stuck). Aligning with our hypotheses, we observed significant improvements in users’ ability to classify the state of the robot after the learning process. In addition, we investigated and confirmed that an algorithm initialization informed by previous user data could be used to speed up the learning process. Finally, we sought to determine if similarities existed between humans when interpreting robot states from nonverbal expressions. The results of our study demonstrated that the method used for algorithm initialization (informed vs. uninformed) strongly influenced whether participants converged to a shared representation. Additionally, we observed that modulating the pitch bend parameter had a significantly greater influence on how users associate sounds with robot states when compared to the other modulated acoustic parameters.

Beyond auditory expressions, our next step is to adapt this approach to function with parameterized expressive motion \cite{Venture2019RobotMotions}. Upon successfully generating parameterized expressive motion, the goal is to integrate these modalities to form a context-adaptive system capable of learning to communicate functional states with multimodal nonverbal expressions.

\bibliographystyle{IEEEtran}

{\footnotesize
\bibliography{references.bib, references_extra.bib}} % Makes the bib text small

\begin{thebibliography}{10}
\providecommand{\url}[1]{#1}
\csname url@rmstyle\endcsname
\providecommand{\newblock}{\relax}
\providecommand{\bibinfo}[2]{#2}
\providecommand\BIBentrySTDinterwordspacing{\spaceskip=0pt\relax}
\providecommand\BIBentryALTinterwordstretchfactor{4}
\providecommand\BIBentryALTinterwordspacing{\spaceskip=\fontdimen2\font plus
\BIBentryALTinterwordstretchfactor\fontdimen3\font minus
  \fontdimen4\font\relax}
\providecommand\BIBforeignlanguage[2]{{%
\expandafter\ifx\csname l@#1\endcsname\relax
\typeout{** WARNING: IEEEtran.bst: No hyphenation pattern has been}%
\typeout{** loaded for the language `#1'. Using the pattern for}%
\typeout{** the default language instead.}%
\else
\language=\csname l@#1\endcsname
\fi
#2}}

\bibitem{Breazeal2005EffectsTeamwork}
C.~Breazeal, C.~D. Kidd, A.~L. Thomaz, G.~Hoffman, and M.~Berlin, ``{Effects of
  nonverbal communication on efficiency and robustness in human-robot
  teamwork},'' \emph{2005 IROS}, 2005.

\bibitem{Knight2016LabanLanguage}
H.~Knight and R.~Simmons, ``{Laban head-motions convey robot state: A call for
  robot body language},'' in \emph{2016 ICRA}, 2016.

\bibitem{Venture2019RobotMotions}
G.~Venture and D.~Kuli{\'{c}}, ``{Robot Expressive Motions},'' \emph{ACM THRI},
  2019.

\bibitem{Dragan2015EffectsCollaboration}
A.~D. Dragan, S.~Bauman, J.~Forlizzi, and S.~S. Srinivasa, ``{Effects of Robot
  Motion on Human-Robot Collaboration},'' \emph{ACM/IEEE International
  Conference on HRI}, 2015.

\bibitem{Baraka2016EnhancingLights}
K.~Baraka, S.~Rosenthal, and M.~Veloso, ``{Enhancing human understanding of a
  mobile robot's state and actions using expressive lights},'' \emph{RO-MAN
  2016}, 2016.

\bibitem{Zahray2020RobotInteraction}
L.~Zahray, R.~Savery, L.~Syrkett, and G.~Weinberg, ``{Robot Gesture
  Sonification to Enhance Awareness of Robot Status and Enjoyment of
  Interaction},'' in \emph{RO-MAN 2020}, 2020.

\bibitem{Frid2022PerceptualGestures}
E.~Frid and R.~Bresin, ``{Perceptual Evaluation of Blended Sonification of
  Mechanical Robot Sounds Produced by Emotionally Expressive Gestures},''
  \emph{International Journal of Social Robotics}, 2022.

\bibitem{FernandezDeGorostizaLuengo2017SoundCues}
J.~Fernandez De Gorostiza~Luengo, F.~Alonso~Martin, A.~Castro-Gonzalez, and
  M.~A. Salichs, ``{Sound synthesis for communicating nonverbal expressive
  cues},'' \emph{IEEE Access}, 2017.

\bibitem{Bellona2017EmpiricallyMovement}
J.~Bellona, L.~Bai, L.~Dahl, and A.~LaViers, ``{Empirically Informed Sound
  Synthesis Application for Enhancing the Perception of Expressive Robotic
  Movement},'' \emph{ICAD 2017}, 2017.

\bibitem{Roy2023TowardsSonification}
L.~Roy, R.~Attfield, D.~Kuli{\'{c}}, and E.~Croft, ``{Towards Improving User
  Experience and Shared Task Performance with Mobile Robots through
  Parameterized Nonverbal State Sonification},'' in \emph{Sound and Robotics:
  Speech, Non-verbal audio and Robotic Musicianship}, 2023.

\bibitem{Zhang2021BringingPerception}
B.~J. Zhang, N.~Stargu, S.~Brimhall, L.~Chan, J.~Fick, and N.~T. Fitter,
  ``{Bringing WALL-E Out of the Silver Screen: Understanding How Transformative
  Robot Sound Affects Human Perception},'' \emph{ICRA}, 2021.

\bibitem{Fernandez2018PassiveInterpretability}
R.~Fernandez, N.~John, S.~Kirmani, J.~Hart, J.~Sinapov, and P.~Stone,
  ``{Passive Demonstrations of Light-Based Robot Signals for Improved Human
  Interpretability},'' \emph{RO-MAN 2018}, 2018.

\bibitem{Zinina2020Non-verbalLikeability}
A.~Zinina, L.~Zaidelman, N.~Arinkin, and A.~Kotov, ``{Non-verbal behavior of
  the robot companion: A contribution to the likeability},'' \emph{Procedia
  Computer Science}, 2020.

\bibitem{SuttonRichardBarto2020ReinforcementLearning}
A.~Sutton, Richard ;~Barto, \emph{{Reinforcement Learning}}, 2020.

\bibitem{Patompak2020LearningInteraction}
P.~Patompak, S.~Jeong, I.~Nilkhamhang, and N.~Y. Chong, ``{Learning Proxemics
  for Personalized Human–Robot Social Interaction},'' \emph{International
  Journal of Social Robotics}, 2020.

\bibitem{McQuillin2022LearningFeedback}
E.~McQuillin, N.~Churamani, and H.~Gunes, ``{Learning Socially Appropriate
  Robo-waiter Behaviours through Real-time User Feedback},'' \emph{ACM/IEEE
  International Conference on HRI}, 2022.

\bibitem{Tseng2018ActiveFeedback}
S.~H. Tseng, F.~C. Liu, and L.~C. Fu, ``{Active Learning on Service Providing
  Model: Adjustment of Robot Behaviors Through Human Feedback},'' \emph{IEEE
  TCDS}, 2018.

\bibitem{Qureshi2016RobotLearning}
A.~H. Qureshi, Y.~Nakamura, Y.~Yoshikawa, and H.~Ishiguro, ``{Robot gains
  social intelligence through multimodal deep reinforcement learning},''
  \emph{IEEE-RAS Humanoids}, 2016.

\bibitem{Akalin2021ReinforcementRobotics}
N.~Akalin and A.~Loutfi, ``{Reinforcement learning approaches in social
  robotics},'' \emph{Sensors (Switzerland)}, 2021.

\bibitem{Dragan2013LegibilityMotion}
A.~D. Dragan, K.~C. Lee, and S.~S. Srinivasa, ``{Legibility and predictability
  of robot motion},'' \emph{ACM/IEEE International Conference on HRI}, 2013.

\bibitem{Churamani2018LearningModulations}
N.~Churamani, P.~Barros, E.~Strahl, and S.~Wermter, ``{Learning Empathy-Driven
  Emotion Expressions using Affective Modulations},'' \emph{IJCNN Proceedings},
  2018.

\bibitem{Leite2012ModellingCompanions}
I.~Leite, G.~Castellano, and S.~Mascarenhas, ``{Modelling Empathy in Social
  Robotic Companions},'' \emph{Advances in User Modeling}, 2012.

\bibitem{Lathuiliere2019NeuralInteraction}
S.~Lathuili{\`{e}}re, B.~Mass{\'{e}}, P.~Mesejo, and R.~Horaud, ``{Neural
  network based reinforcement learning for audio–visual gaze control in
  human–robot interaction},'' \emph{Pattern Recognition Letters}, 2019.

\bibitem{Papaioannou2017HybridLearning}
I.~Papaioannou, C.~Dondrup, J.~Novikova, and O.~Lemon, ``{Hybrid chat and task
  dialogue for more engaging HRI using reinforcement learning},'' \emph{RO-MAN
  2017}, 2017.

\bibitem{Slivkins2019IntroductionBandits}
A.~Slivkins, ``{Introduction to multi-armed bandits},'' \emph{Foundations and
  Trends in Machine Learning}, 2019.

\bibitem{Kuleshov2014AlgorithmsProblems}
V.~Kuleshov and D.~Precup, ``{Algorithms for multi-armed bandit problems},''
  \emph{Journal of Machine Learning Research}, 2014.

\bibitem{Lattimore2020BanditAlgorithms}
T.~Lattimore and C.~Szepesv{\'{a}}ri, ``{Bandit Algorithms},'' 2020.

\bibitem{Wilde2021LearningFeedback}
N.~Wilde, E.~Bıyık, D.~Sadigh, and S.~L. Smith, ``{Learning Reward Functions
  from Scale Feedback},'' in \emph{CoRL}, 2021.

\bibitem{Gajos2017}
K.~Z. Gajos and K.~Chauncey, ``{The influence of personality traits and
  cognitive load on the use of adaptive user interfaces},'' \emph{IUI
  Proceedings}, 2017.

\bibitem{Sekmen2013AssessmentInteractions}
A.~Sekmen and P.~Challa, ``{Assessment of adaptive human-robot interactions},''
  \emph{Knowledge-Based Systems}, 2013.

\bibitem{Gajos2010AutomaticallySupple}
K.~Z. Gajos, D.~S. Weld, and J.~O. Wobbrock, ``{Automatically generating
  personalized user interfaces with Supple},'' \emph{Artificial Intelligence},
  2010.

\bibitem{Ghadirzadeh2021Human-CenteredLearning}
A.~Ghadirzadeh, X.~Chen, W.~Yin, Z.~Yi, M.~Bjorkman, and D.~Kragic,
  ``{Human-Centered Collaborative Robots with Deep Reinforcement Learning},''
  \emph{IEEE RA-L}, 2021.

\bibitem{Weber2018HowLearning}
K.~Weber, H.~Ritschel, I.~Aslan, F.~Lingenfelser, and E.~Andr{\'{e}}, ``{How to
  shape the humor of a robot - Social behavior adaptation based on
  reinforcement learning},'' \emph{ICMI Proceedings}, 2018.

\end{thebibliography}
% \bibliography{references.bib}

% mic drop
\end{document}